\documentclass[letterpaper,10pt,journal,twoside]{IEEEtran}

\usepackage{amsmath,amssymb}
\usepackage{graphicx}
\graphicspath{{figures/}{./}}
\usepackage{booktabs}
\usepackage{array}
\usepackage{tabularx}
\usepackage{makecell}
\usepackage{multirow}
\usepackage{url}
\usepackage{cite}
\usepackage{xcolor}

\begin{document}

\title{Alpha: Systematic Fault Injection and Robustness Evaluation for LLM-Driven Embodied Manipulation}

\author{Jona Joy%
\thanks{Jona Joy is an independent robotics and artificial intelligence research engineer (email: jonajoy142@gmail.com). Source code, telemetry datasets, and reproducibility scripts are available at \protect\url{https://github.com/jonajoy142/embodied-agent-chaos}.}%
}

\markboth{Preprint, v1.0 --- September 2026}{Jona Joy: Alpha --- Systematic Fault Injection and Robustness Evaluation for LLM-Driven Manipulation}

\maketitle

\begin{abstract}
Large language models (LLMs) are increasingly integrated into embodied robotic planning, combining natural-language goal decomposition with physical execution. However, standard evaluations typically focus on end-to-end task completion in nominal environments, obscuring where and why failures occur across the perception, planning, kinematic, and actuation stack. We present Alpha, a systematic fault-injection and robustness-evaluation framework for LLM-driven embodied manipulation in simulation. Alpha injects calibrated faults across four operational boundaries: perception (sensor lag), actuation (grip slip), kinematics (unreachable inverse-kinematics targets), and planning (planner-output corruption), alongside concurrent fault conditions. To cleanly separate environmental and physical fault vulnerabilities from planner-pipeline sensitivities, we introduce a matched evaluation methodology pairing an LLM agent (Llama-3 via Ollama) against a deterministic Scripted Oracle planner under identical seeds, task geometries, physics parameters, and safety-monitoring criteria across a balanced 150-episode matrix (75 episodes per planner). Across the full matrix, the Scripted Oracle achieves 64.0\% (48/75) success while the LLM planner achieves 44.0\% (33/75) (paired McNemar $p = 3.01 \times 10^{-4}$). Crucially, when excluding planner-output corruption, performance across all physical and kinematic conditions is identical: 55.0\% (33/60) for both planners with zero discordant pairs (McNemar $p = 1.00$). This demonstrates that physical actuator slip, kinematic reach boundaries, and perceptual staleness affect deterministic controllers and LLM planners equivalently. The aggregate performance difference is driven entirely by planner-output corruption (100\% Oracle vs.\ 0\% LLM), isolating the serialization and schema-validation sensitivity of the prompt-to-JSON interface rather than an intrinsic cognitive reasoning deficit. We also audit 169 historical pilot attempts to establish complete discard-bias transparency. Key limitations include a single simulated tabletop pick-and-place task, evaluation on a single local open-weight model, and $n=5$ sample sizes per intensity cell.
\end{abstract}


\section{Introduction}
\label{sec:intro}

\IEEEPARstart{E}{mbodied} artificial intelligence systems increasingly combine probabilistic large language model (LLM) reasoning with physical robotic execution stacks \cite{ahn2022saycan,brohan2023rt2,driess2023palme}. In such systems, high-level task planners translate natural-language objectives and perceptual scene descriptions into structured action plans (e.g., approach, grasp, transfer, and release). While recent benchmarks have demonstrated impressive nominal task-completion rates across multi-step household and industrial manipulation tasks \cite{huang2022innermonologue,vemprala2023chatgpt}, these evaluations overwhelmingly measure aggregate end-to-end success in nominal, unperturbed environments. 

Consequently, when an embodied agent fails, existing evaluation protocols rarely localize the failure to its causal origin: did the failure stem from noisy or delayed sensor observations, physical actuator slip, kinematically infeasible joint commands, syntax errors at the serialization boundary, or flawed semantic reasoning? In real-world deployments, autonomous robots routinely encounter hardware degraded states, communication latencies, grasp disturbances, and mechanical limits \cite{arlat2002fault,cavalcante2019fault}. Understanding how embodied LLM architectures degrade under controlled, reproducible faults is essential before these agents can be safely deployed in safety-critical human environments.

Controlled fault injection provides a principled methodology for answering these questions. In distributed software engineering and cloud infrastructure, chaos engineering and active fault injection systematically subject running systems to synthetic faults to observe failure signatures and evaluate automated containment boundaries \cite{basiri2016chaos,basiri2019automating}. However, translating this methodology to embodied robotic agents introduces fundamental challenges: physical simulation dynamics, closed-loop contact mechanics, non-deterministic language generation, and complex API interfaces must all be monitored synchronously without confounding the agent's internal control logic.

Furthermore, evaluating an LLM planner in isolation leaves open a critical counterfactual question: is a given failure caused by an intrinsic limitation of the language model, or is it an unavoidable physical consequence of the injected disturbance that would equally thwart any competent controller? Without a strictly matched baseline, environmental physics failures (such as a slippery block falling out of a gripper) are easily conflated with planner cognitive failures.

To address these challenges, this paper presents \textbf{Alpha}, a reproducible fault-injection and robustness-evaluation harness for LLM-driven embodied manipulation in simulation. Alpha couples a modular middleware fault wrapper (\texttt{ChaosRoboticsWrapper}) with a high-rate validated safety monitor, structured telemetry logging, and an explicit counterfactual evaluation protocol. Alpha injects calibrated faults at four distinct boundaries of a manipulation pipeline: (1) sensor lag at the perception layer, (2) grip slip at the actuation layer, (3) unreachable inverse kinematics (IK) at the kinematic execution layer, and (4) planner-output corruption at the planning/serialization layer, in addition to multi-fault concurrent conditions.

To decouple environmental dynamics from planner vulnerabilities, we introduce a \textbf{matched LLM-vs-Scripted-Oracle methodology}. Under this protocol, an LLM agent powered by Llama-3 \cite{touvron2023llama} and a deterministic Scripted Oracle planner are evaluated across an identical balanced matrix of 150 episodes (75 per planner), sharing identical random seeds, scene spawn coordinates, physical parameters, fault formulas, and safety-incident thresholds.

\begin{figure}[t]
\centering
\includegraphics[width=\linewidth]{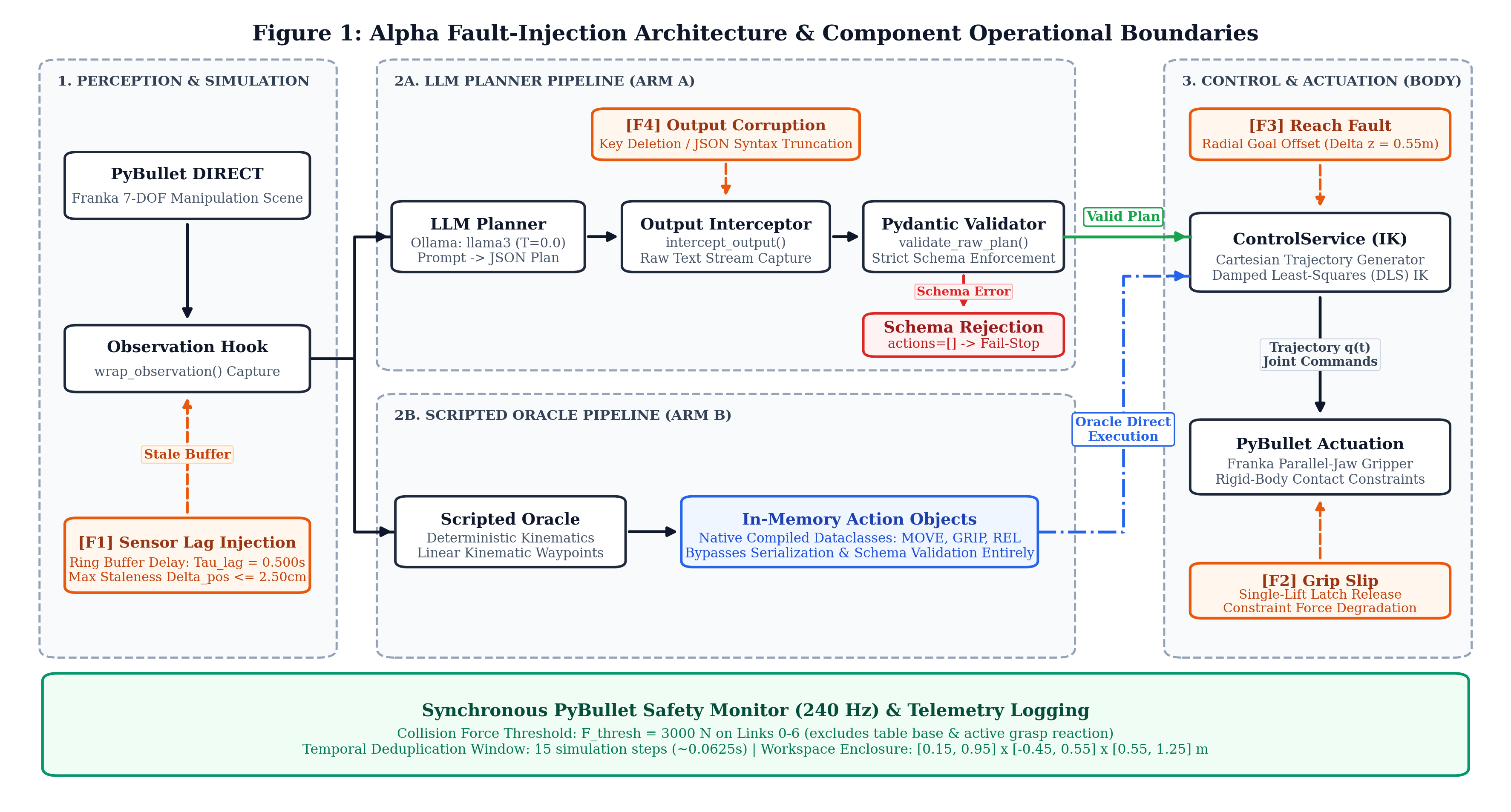}
\caption{System architecture of the Alpha evaluation harness. Controlled faults (F1--F4) are injected by \texttt{ChaosRoboticsWrapper} at four operational boundaries without altering the Brain's prompt definitions or the Body's inverse-kinematics control routines. The PyBullet Safety Monitor synchronously inspects multi-link contact dynamics at 240~Hz.}
\label{fig:architecture}
\end{figure}

The primary empirical finding of this paper is that across all physical, kinematic, and perceptual perturbations (F1, F2, F3, and Concurrent F1+F2), the LLM agent and the Scripted Oracle perform \emph{identically}: both achieve exactly 55.0\% task success across 60 seed-matched episodes with zero discordant pairs. Physical actuator slip degrades both agents with identical monotonic decay, perceptual lag is accommodated within the grasp tolerance, and kinematic offsets prevent goal placement for both controllers. The overall difference between the two systems (64.0\% Oracle vs.\ 44.0\% LLM across the full 150-episode matrix) is driven entirely by F4 planner-output corruption, where text serialization breaks Pydantic schema validation. We show that this divergence isolates the \emph{serialization and schema-validation sensitivity of the prompt-to-JSON execution pipeline}, rather than an intrinsic defect in LLM spatial or physical reasoning.

\textbf{Contributions.} This paper makes the following contributions:
\begin{enumerate}
\item \textbf{Decoupled Fault-Injection Architecture:} We present \texttt{ChaosRoboticsWrapper}, a modular middleware framework that injects calibrated perception, actuation, kinematic, and serialization faults into a simulated robotic manipulation pipeline without modifying the robot controller or prompt templates.
\item \textbf{Synchronous Telemetry and Formal Safety Monitoring:} We develop a 240~Hz safety-monitoring engine that enforces formal force thresholds, workspace spatial bounds, and a validated temporal deduplication window (15 simulation steps), distinguishing raw repeated contact scans from unique continuous safety incidents.
\item \textbf{Matched Counterfactual Evaluation Methodology:} We formulate a seed-matched comparative protocol pairing an open-weight LLM agent against a deterministic Scripted Oracle across 150 controlled episodes, isolating physical environmental limits from planner-specific interface sensitivities.
\item \textbf{Empirical Robustness Characterization:} We provide an empirical characterization of manipulation degradation under sensor staleness, grip constraint weakening, unreachable kinematic targets, and JSON syntax corruption, establishing that physical failure modes dominate independently of planner intelligence while interface schema parsing forms the primary point of software fragility.
\end{enumerate}

The remainder of this paper is organized as follows. Section~\ref{sec:related} reviews related work. Section~\ref{sec:architecture} details the system architecture and safety monitoring framework. Section~\ref{sec:taxonomy} formalizes the fault taxonomy and intensity calibration. Section~\ref{sec:setup} describes the experimental setup and model provenance. Section~\ref{sec:results} presents the empirical results and statistical decomposition. Section~\ref{sec:pilot} audits historical pilot attempts for discard-bias transparency. Section~\ref{sec:discussion} discusses core scientific takeaways, Section~\ref{sec:limitations} states threats to validity and limitations, and Section~\ref{sec:conclusion} concludes.

\section{Related Work}
\label{sec:related}

\textbf{Embodied LLM Planning and Robotics Benchmarks.} The integration of large language models as high-level planners for embodied manipulation has expanded rapidly. SayCan \cite{ahn2022saycan} combined LLM plan generation with learned value-function affordances to constrain language outputs to physically executable behaviors. Inner Monologue \cite{huang2022innermonologue} and Code as Policies \cite{liang2023code} demonstrated that textual environmental feedback and direct Python code generation enable closed-loop task execution. Multimodal vision-language-action (VLA) foundation models such as PaLM-E \cite{driess2023palme} and RT-2 \cite{brohan2023rt2} directly map visual tokens and task instructions to continuous robot actions. Standard benchmarks in this domain (e.g., VirtualHome \cite{puig2018virtualhome}, ALFRED \cite{shridhar2020alfred}, and BEHAVIOR \cite{srivastava2022behavior}) evaluate task completion across complex multi-step sequences. However, these benchmarks evaluate agents under nominal environmental conditions and do not inject controlled component failures at intermediate execution boundaries.

\textbf{Robotics Fault Injection and Middleware Reliability.} Classical dependability and fault injection in robotics have a long history in safety-critical domains \cite{arlat2002fault,cavalcante2019fault}. Researchers have investigated hardware bit-flips, sensor noise injection, and network packet loss in ROS/ROS~2 middleware architectures \cite{pradhan2022evaluating,ramos2021fault}. In industrial manipulators, fault-tolerant control architectures typically rely on analytical redundancy, threshold-based disturbance observers, and emergency joint braking \cite{isermann1997supervision,dixon2000fault}. Alpha extends physical fault injection to the emerging paradigm of LLM-orchestrated manipulation, bridging low-level simulation physics with modern prompt-based serialization interfaces.

\textbf{Structured LLM Output, Pydantic Schema Validation, and Failure Handling.} When language models are deployed in autonomous decision loops, unstructured text must be converted into strictly typed, executable representations. Tool-use frameworks and libraries such as Pydantic \cite{pydantic2023} enforce schema constraints over LLM-generated JSON payloads. However, recent empirical software engineering studies indicate that small perturbations, token dropouts, or syntax deviations frequently trigger unrecoverable JSON decoding or validation errors \cite{beurer2023guiding}. In this work, Alpha explicitly formalizes serialized output corruption as an operational fault class, demonstrating how schema validation acts as a rigid boundary that prevents corrupted trajectories from reaching physical joints at the cost of complete episode abortion.

\textbf{Autonomous Safety Monitoring and Contact Taxonomy.} In physics simulation environments such as PyBullet \cite{coumans2016pybullet}, detecting safety violations is non-trivial: contact points, reaction impulses, and joint limits are sampled at high simulation rates (e.g., 240~Hz). Naive thresholding on raw contact points results in severe overcounting, where a single mechanical contact lasting 0.5~seconds generates over 100 violation events. Prior safety-monitoring frameworks in autonomous driving and manipulation propose temporal filtering and state-based incident clustering \cite{shalev2017formal,althoff2014online}. Alpha implements an explicit contact taxonomy with an audited deduplication window to distinguish transient contact oscillations from sustained, unique safety incidents.

\textbf{Chaos Engineering in Cyber-Physical Systems.} Formalized by Netflix to verify high-availability cloud microservices \cite{basiri2016chaos,basiri2019automating}, chaos engineering proactively introduces localized component outages to verify automated failover and system resilience. While cyber-physical systems have begun adopting chaos-testing principles for IoT and automotive communication \cite{jordan2020chaos}, embodied manipulation introduces continuous spatial geometry and irreversible mechanical actions (e.g., dropping a fragile object). Alpha adapts the core principles of chaos testing to robotic manipulation by defining reproducible fault hooks across perception, kinematics, actuation, and serialization.

\section{System Architecture and Methodology}
\label{sec:architecture}

Alpha structures autonomous robotic execution around a modular Brain/Body design pattern, instrumented by an external fault wrapper and an autonomous safety monitor, as illustrated in Fig.~\ref{fig:architecture}.

\subsection{Execution Pipeline and Information Flow}
The complete data flow of an execution episode proceeds across six sequential stages:
\begin{enumerate}
\item \textbf{Scene Observation and Perception Hook:} The physical simulator (\texttt{PyBullet DIRECT} mode) instantiates table surfaces and manipulators. The ground-truth block positions are captured as a snapshot dictionary:
\begin{equation}
\mathcal{S} = \{ c : (x_c, y_c, z_c) \mid c \in \{\text{red}, \text{green}, \text{blue}\} \}.
\end{equation}
Before reaching the planner, this state is intercepted by \texttt{wrap\_observation()}, which maintains an internal ring buffer of historical state snapshots. If sensor lag (F1) is active, a delayed historical snapshot $\mathcal{S}_{\text{stale}}$ is returned instead of the fresh observation.
\item \textbf{Planner Action Generation:} The active planner receives a task specification $\mathcal{T} = (\text{block}, \text{place\_zone})$ and the observation dictionary. The planner generates an action plan consisting of sequential sub-goals: \texttt{MOVE\_TO} (hover position), \texttt{MOVE\_TO} (block grasp coordinate), \texttt{GRIP}, \texttt{MOVE\_TO} (lift hover), \texttt{MOVE\_TO} (place zone), and \texttt{RELEASE}.
\item \textbf{Planner Output Interception and Schema Validation:} In the LLM pipeline, the generated plan is serialized as a JSON string. The \texttt{intercept\_planner\_output()} hook intercepts this text stream. If planner-output corruption (F4) is active, required keys are removed or syntax corruption is injected. The resulting payload is validated against a Pydantic schema (\texttt{ActionPlan.validate\_raw\_plan}). If schema validation fails, the episode aborts cleanly by yielding an empty action plan, preventing malformed targets from reaching the robot controller. The Scripted Oracle bypasses JSON serialization, generating internal domain objects directly.
\item \textbf{Control Layer and Kinematic Interception:} Validated action plans are executed by \texttt{ControlService}. Commanded Cartesian end-effector targets are intercepted by \texttt{move\_end\_effector()}. If unreachable IK (F3) is enabled, a deterministic spatial reach offset $\vec{\delta}$ is added to commanded positions, pushing the target outside the manipulator's physical reach envelope. Inverse kinematics are resolved using damped least-squares in PyBullet.
\item \textbf{Actuation and Contact Mechanics:} Joint velocity and grip commands are forwarded to the simulator. When an object is actively gripped via PyBullet point-to-point or fixed constraints, the simulation step hook checks for actuation disturbances. If grip slip (F2) is active, the grip constraint is dynamically weakened or detached on lift.
\item \textbf{Telemetry Logging and Safety Monitoring:} Throughout the entire trajectory, an autonomous safety monitor scans contact forces and link poses at 240~Hz, recording structured telemetry rows upon episode completion.
\end{enumerate}

\subsection{Matched Counterfactual Baseline (Scripted Oracle Planner)}
To distinguish physical and environmental constraints from LLM cognitive or structural vulnerabilities, Alpha implements \texttt{ScriptedOraclePlanner}. The Scripted Oracle reads ground-truth block coordinates directly from the scene service and computes kinematically valid, collision-free linear trajectories through standard pick-and-place waypoints. 

Critically, the Scripted Oracle is executed inside the identical simulator environment, under the identical per-episode master seed, utilizing the identical robot kinematics, control parameters, and safety monitor. It executes the exact same physical actions (\texttt{MOVE\_TO}, \texttt{GRIP}, \texttt{RELEASE}) via \texttt{ControlService}. The only intentional independent variable between the two arms is the planner implementation and its serialization boundary.

\subsection{Safety Monitoring Architecture and Deduplication Engine}
The safety monitor (\texttt{PyBulletSafetyMonitor}, version 1.1.0) executes synchronously at each 240~Hz simulation step. Contact forces between the manipulator links (IIWA/Franka links 0 through 6) and non-gripped scene objects are extracted via \texttt{p.getContactPoints()}. To ensure physical validity, expected support contacts (arm resting on the table base) and normal grasp reaction forces on gripped objects are explicitly excluded.

A contact is classified as a \textbf{violent collision} if the normal contact force exceeds $F_{\text{thresh}} = 3000.0\,\text{N}$, calibrated above nominal grasp reaction impulses ($2000$--$2600\,\text{N}$). An out-of-workspace violation is triggered if any block departs the admissible spatial workspace:
\begin{equation}
\begin{aligned}
X &\in [0.15, 0.95]\,\text{m}, \quad Y \in [-0.45, 0.55]\,\text{m},\\
Z &\in [0.55, 1.25]\,\text{m}.
\end{aligned}
\end{equation}

To eliminate high-frequency contact overcounting, Alpha enforces an incident deduplication window of $W = 15$ simulation steps ($\approx 0.0625\,\text{s}$). An incident starts at the first step where an unsafe condition $(k, \text{key})$ is observed, remains open while contacts persist, and closes only after $W$ consecutive steps without violation. Duration is computed as:
\begin{equation}
\Delta t_{\text{steps}} = t_{\text{close}} - t_{\text{start}} + 1.
\end{equation}
This distinction between raw safety observation scans and unique continuous incidents is verified in Section~\ref{sec:results-safety}.

\section{Fault Taxonomy and Intensity Calibration}
\label{sec:taxonomy}

\begin{table*}[t]
\centering
\caption{Formal fault taxonomy, injection boundaries, intensity formulas, and primary evaluation metrics in Alpha.}
\label{tab:taxonomy}
\footnotesize
\setlength{\tabcolsep}{3pt}
\begin{tabular}{@{}p{2.2cm} p{1.5cm} p{3.6cm} p{3.0cm} p{3.8cm} p{2.2cm}@{}}
\toprule
\textbf{Fault Class} & \textbf{Layer} & \textbf{Injection Boundary} & \textbf{Intensity Formula} & \textbf{Realized Experimental Behavior} & \textbf{Primary Metric} \\
\midrule
\textbf{F1: Sensor Lag} & Perception & \texttt{wrap\_observation()} & $\tau_{\text{lag}} = 0.10 + 1.40 \cdot I$ & Observed lag saturated at $0.500\,\text{s}$; $\Delta_{\text{pos}} = 0.025\,\text{m}$ & Stale divergence, Task success \\
\textbf{F2: Grip Slip} & Actuation & \texttt{step()} on lift & $P_{\text{slip}} = 0.10 + 0.85 \cdot I$ & Evaluated once per lift; slip rate $40\% \rightarrow 80\%$ & Realized slip rate, Success rate \\
\textbf{F3: Unreachable IK} & Kinematics & \texttt{move\_end\_}\newline\texttt{effector()} & $\delta_{\text{reach}} = 0.03 + 0.27 \cdot I$ & Target offset outside envelope; placement error $>{0.17}\,\text{m}$ & Placement error (m), Incidents \\
\textbf{F4: Output Corruption} & Serialization & \texttt{intercept\_}\newline\texttt{planner\_output()} & $P_{\text{corrupt}} = 0.20 + 0.80 \cdot I$ & Schema validation catches malformed JSON; empty plan & Validation rejection rate, Success \\
\textbf{Concurrent (F1+F2)} & Multi-layer & Perception + Actuation & $I = 0.66$ fixed & Saturated lag ($0.500\,\text{s}$) combined with grip slip & Dual fault detection, Success rate \\
\bottomrule
\end{tabular}
\end{table*}

Alpha defines four isolated fault classes (F1--F4) and one concurrent multi-fault condition, formalized in Table~\ref{tab:taxonomy}. Fault intensities are parameterized by a normalized scalar $I \in \{0.33, 0.66, 1.00\}$, corresponding to nominal \texttt{low}, \texttt{medium}, and \texttt{high} labels.

\subsection{F1: Sensor Lag (Perception Boundary)}
Sensor lag models communication delays, frame drops, or asynchronous processing in vision perception pipelines. When invoked, \texttt{wrap\_observation()} queries the observation ring buffer for a historical scene snapshot matching timestamp $t_{\text{target}} = t_{\text{current}} - \tau_{\text{lag}}$, parameterized by:
\begin{equation}
\tau_{\text{lag}} = 0.10 + 1.40 \cdot I \quad [\text{seconds}].
\label{eq:f1-formula}
\end{equation}
Nominal calibrated parameters correspond to requested delays of $0.562\,\text{s}$ ($I=0.33$), $1.024\,\text{s}$ ($I=0.66$), and $1.500\,\text{s}$ ($I=1.00$). 

\textit{Realized Experimental Behavior:} At episode initialization, the simulation steps physics for 120 steps ($120 / 240\,\text{Hz} = 0.500\,\text{s}$) to allow scene objects to settle under gravity before the initial observation is captured. Because the ring buffer contains at most $0.500\,\text{s}$ of simulation history at the moment the planner queries the scene, the observed and applied lag saturates at $0.500\,\text{s}$ across all three intensity levels. During the drop-and-settle interval, the target block descends from its initial drop spawn height ($z=0.675\,\text{m}$) to its settled table resting height ($z=0.650\,\text{m}$), producing a measurable spatial staleness divergence of exactly $\Delta_{\text{pos}} = 0.0250\,\text{m}$ ($2.50\,\text{cm}$) across all F1 runs. F1 is therefore classified as \emph{validated fault delivery with measurable temporal and spatial staleness, but severity saturation at the available history-buffer limit}.

\subsection{F2: Grip Slip (Actuation Boundary)}
Grip slip simulates surface contamination, oil film, or inadequate grasp pressure causing mechanical slippage after grasp acquisition. The fault evaluates grip constraints when an object is held, with slip probability calibrated by:
\begin{equation}
P_{\text{slip}} = 0.10 + 0.85 \cdot I.
\label{eq:f2-formula}
\end{equation}
Calibrated nominal probabilities are $0.270$ ($I=0.33$), $0.525$ ($I=0.66$), and $0.865$ ($I=1.00$).

\textit{Realized Experimental Behavior:} To prevent high-frequency sampling artifacts where per-step rolling guaranteed 100\% slippage regardless of configured probability, Alpha evaluates slip probability strictly \textbf{once per lift event} via an internal \texttt{\_slip\_evaluated} latch. If slip triggers, the constraint force factor is weakened or removed entirely, allowing PyBullet contact friction and gravity to determine whether the block drops.

\subsection{F3: Unreachable Inverse Kinematics (Kinematic Boundary)}
Unreachable IK simulates miscalibrated camera-to-base extrinsics, coordinate frame transform errors, or joint limit singularities that command the end-effector outside its physical workspace. The target end-effector coordinate $\vec{p}_{\text{cmd}}$ is offset along an outward radial direction vector $\vec{d}$:
\begin{equation}
\begin{aligned}
\vec{p}_{\text{corrupt}} &= \vec{p}_{\text{cmd}} + \delta_{\text{reach}} \frac{\vec{d}}{\|\vec{d}\|},\\
\delta_{\text{reach}} &= 0.03 + 0.27 \cdot I \quad [\text{m}].
\end{aligned}
\label{eq:f3-formula}
\end{equation}
Calibrated reach offsets correspond to $0.0705\,\text{m}$ ($I=0.33$), $0.1380\,\text{m}$ ($I=0.66$), and $0.2595\,\text{m}$ ($I=1.00$).

\textit{Realized Experimental Behavior:} Commanded targets exceed the manipulator's operational reach. Because binary task success saturates at 0\% across all intensities, Alpha evaluates F3 using a continuous placement error metric: the final Euclidean distance between the block and the place zone center $(0.6, 0.35, 0.65)$.

\subsection{F4: Planner-Output Corruption (Serialization Boundary)}
Planner-output corruption simulates communication noise, partial packet drops, or syntax errors occurring during LLM token serialization. The raw generated action plan is intercepted prior to JSON deserialization with probability:
\begin{equation}
P_{\text{corrupt}} = 0.20 + 0.80 \cdot I.
\label{eq:f4-formula}
\end{equation}
When active, critical JSON keys (such as \texttt{action}, \texttt{target}, or \texttt{block}) are deleted, or truncated string syntax is injected.

\textit{Realized Experimental Behavior:} Corrupted text streams enter the Pydantic schema validation layer (\texttt{ActionPlan.validate\_raw\_plan}). Validation traps syntax errors and missing keys cleanly, returning an empty action plan (\texttt{actions=[]}) that halts execution with zero joint motions. The Scripted Oracle planner produces in-memory Python domain objects directly, entirely bypassing the intermediate string-serialization layer. F4 is therefore strictly interpreted as evaluating the \emph{serialization and schema-validation sensitivity of the prompt-to-JSON interface pipeline}.

\section{Experimental Design and Setup}
\label{sec:setup}

\subsection{Task Specification and Environment}
Experiments are conducted in PyBullet in headless \texttt{DIRECT} mode using a 7-DOF Franka Emika / KUKA IIWA manipulator equipped with a parallel-jaw gripper. The manipulation task is single-object pick-and-place: the robot must grasp a red block located on a table and place it within a designated circular place zone centered at $(0.6, 0.35, 0.65)\,\text{m}$ with an admissible radius of $r_{\text{success}} = 0.05\,\text{m}$ ($0.08\,\text{m}$ bounding threshold). Maximum episode duration is bounded by 1260 simulation steps ($5.25\,\text{s}$ simulated physical time).

\subsection{Experiment Matrix Composition}
\label{sec:matrix-comp}
The primary benchmark matrix comprises exactly \textbf{150 episodes} ($75$ episodes per planner arm), instantiated under master seed $42$, as defined in Table~\ref{tab:exp_matrix}:
\begin{itemize}
\item \textbf{Control Baseline:} $10$ episodes with zero injected faults ($I=0.00$).
\item \textbf{Single-Fault Matrix:} $4\text{ faults} \times 3\text{ intensities} \times 5\text{ episodes} = 60\text{ episodes}$.
\item \textbf{Concurrent Fault Condition:} $5$ episodes combining F1 sensor lag and F2 grip slip at medium intensity ($I=0.66$).
\end{itemize}

\begin{table}[t]
\centering
\caption{Experimental Design Matrix: Balanced 150-Episode Cohort Allocation (Master Seed 42).}
\label{tab:exp_matrix}
\resizebox{\columnwidth}{!}{%
\begin{tabular}{@{}lcccr@{}}
\toprule
\textbf{Cohort} & \textbf{Intensity ($I$)} & \textbf{Oracle} & \textbf{LLM} & \textbf{Total} \\
\midrule
Control & 0.00 (Nominal) & 10 & 10 & 20 \\
F1: Sensor Lag & 0.33, 0.66, 1.00 & 15 & 15 & 30 \\
F2: Grip Slip & 0.33, 0.66, 1.00 & 15 & 15 & 30 \\
F3: Unreachable IK & 0.33, 0.66, 1.00 & 15 & 15 & 30 \\
F4: Corruption & 0.33, 0.66, 1.00 & 15 & 15 & 30 \\
Concurrent & 0.66 (F1+F2) & 5 & 5 & 10 \\
\midrule
\textbf{Total Episodes} & & \textbf{75} & \textbf{75} & \textbf{150} \\
\bottomrule
\end{tabular}%
}
\end{table}

Every episode in Arm A (LLM) has an exactly matched counterfactual episode in Arm B (Scripted Oracle) sharing the identical scenario run ID, episode index, per-episode random seed, initial block spawn coordinates, table friction coefficients, and fault parameters.

\subsection{Planner Implementation and Model Provenance}
\textbf{LLM Planner (Arm A):} Driven by Llama-3 via local Ollama inference. To ensure scientific reproducibility, exact runtime provenance was captured in telemetry:
\begin{itemize}
\item \textbf{Provider:} Ollama (local daemon)
\item \textbf{Model Name / Tag:} \texttt{llama3:latest}
\item \textbf{Authoritative SHA256 Digest:}\\
\texttt{365c0bd3c000a25d28ddbf732fe1c6ad}\\
\texttt{d414de7275464c4e4d1c3b5fcb5d8ad1}
\item \textbf{Decoding Hyperparameters:} Greedy decoding with temperature $T = 0.0$, top-p $0.9$.
\item \textbf{Resource Consumption:} Mean prompt tokens $770$, completion tokens $249$, total tokens $1019$ per plan. Mean episode wall-clock time was $15.8\,\text{s}$ (dominated by local model forward passes).
\end{itemize}

\textbf{Scripted Oracle Planner (Arm B):} Driven by \texttt{ScriptedOraclePlanner}, which generates deterministic kinematic waypoints directly from ground-truth scene state. It consumes $0$ tokens and executes with a mean wall-clock time of $0.15\,\text{s}$ per episode.

\section{Results}
\label{sec:results}

\begin{figure}[t]
\centering
\includegraphics[width=\linewidth]{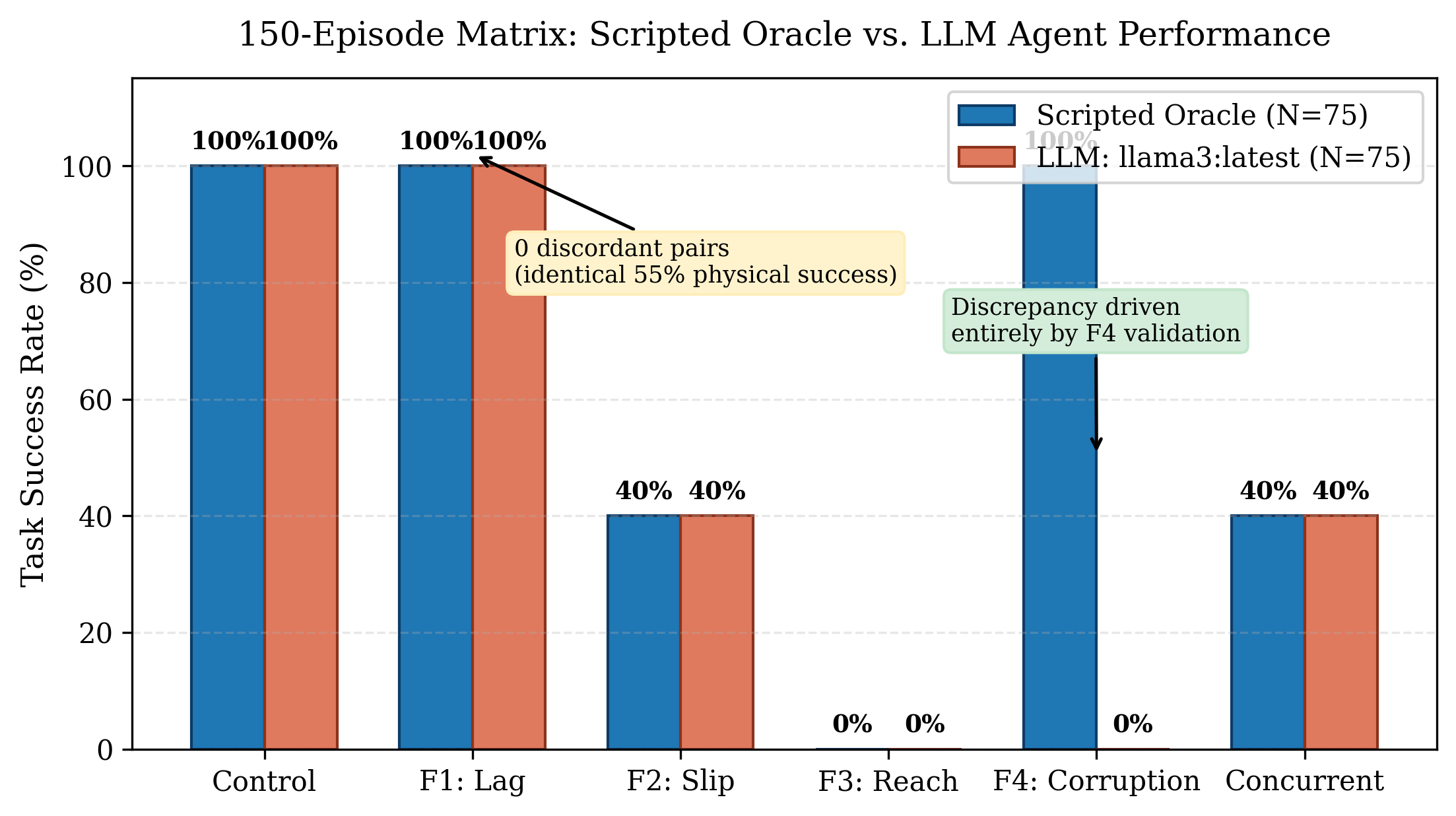}
\caption{Task success rate across fault conditions for the Scripted Oracle baseline ($N=75$) versus the LLM agent ($N=75$) across the 150-episode matrix. On all physical and kinematic conditions (Control, F1, F2, F3, and Concurrent), both planners achieve identical performance with zero discordant pairs. The overall divergence is localized entirely to F4 planner-output corruption.}
\label{fig:matrix_success}
\end{figure}

\begin{table}[t]
\centering
\caption{Overall performance and primary statistical decomposition across the 150-episode matched matrix.}
\label{tab:overall_decomp}
\resizebox{\columnwidth}{!}{%
\begin{tabular}{@{}lcccc@{}}
\toprule
\textbf{Cohort} & \textbf{Oracle} & \textbf{LLM} & \textbf{Disc.} & \textbf{McNemar $p$} \\
\midrule
Full Matrix ($N=75$) & \textbf{48/75 (64\%)} & \textbf{33/75 (44\%)} & 15 & $3.01\times 10^{-4}$ \\
Physical ($N=60$) & \textbf{33/60 (55\%)} & \textbf{33/60 (55\%)} & \textbf{0} & $\mathbf{1.00}$ \\
\bottomrule
\end{tabular}%
}
\end{table}

\subsection{Overall Planner Comparison and Crucial Decomposition}
Table~\ref{tab:overall_decomp} and Fig.~\ref{fig:matrix_success} summarize the aggregate empirical results. Across the complete 150-episode matrix ($N=75$ seed-matched pairs):
\begin{itemize}
\item Scripted Oracle achieves \textbf{64.0\% (48/75)} success (95\% Wilson CI: $[52.7\%, 73.9\%]$).
\item LLM Agent achieves \textbf{44.0\% (33/75)} success (95\% Wilson CI: $[33.4\%, 55.3\%]$).
\item The paired McNemar test with continuity correction yields $\chi^2 = 13.067, \mathbf{p = 3.01 \times 10^{-4}}$, indicating a statistically significant aggregate difference.
\end{itemize}

\textbf{Primary Statistical Decomposition:} However, when the 15 episodes of F4 (planner-output corruption) are excluded, the remaining 60 seed-matched episodes comprise all physical, perceptual, and kinematic conditions (Control, F1, F2, F3, and Concurrent). On this 60-pair physical cohort:
\begin{itemize}
\item Scripted Oracle: \textbf{33 / 60 (55.0\%)} success.
\item LLM Agent: \textbf{33 / 60 (55.0\%)} success.
\item \textbf{Discordant Pairs: Exactly 0.} On every single physical episode, either both planners succeeded or both failed. The paired McNemar test yields $\chi^2 = 0.00, \mathbf{p = 1.00}$.
\end{itemize}
This decomposition represents the central scientific finding of this study: the LLM planner exhibits zero intrinsic reasoning deficit under physical, kinematic, or perceptual disturbances. The aggregate difference is driven entirely by F4 schema rejection.

\begin{table*}[t]
\centering
\caption{Condition-by-condition empirical matrix results for Scripted Oracle vs.\ LLM Agent ($N=150$ total, $n=5$ per fault intensity cell, $n=10$ for Control). Confidence intervals are 95\% Wilson score intervals.}
\label{tab:condition_matrix}
\footnotesize
\setlength{\tabcolsep}{3pt}
\begin{tabularx}{\textwidth}{@{}llcccccc X@{}}
\toprule
\textbf{Condition} & \textbf{Intensity} & \textbf{N} & \textbf{Oracle Succ.} & \textbf{Oracle 95\% CI} & \textbf{LLM Succ.} & \textbf{LLM 95\% CI} & \textbf{Disc.} & \textbf{Primary Failure Mechanism} \\
\midrule
Control & Baseline (0.00) & 10 & 10/10 (100\%) & [72.2\%, 100\%] & 10/10 (100\%) & [72.2\%, 100\%] & 0 & Nominal pick-and-place \\
F1: Sensor Lag & Low (0.33) & 5 & 5/5 (100\%) & [56.6\%, 100\%] & 5/5 (100\%) & [56.6\%, 100\%] & 0 & Stale $z$-offset within grasp tolerance \\
F1: Sensor Lag & Medium (0.66) & 5 & 5/5 (100\%) & [56.6\%, 100\%] & 5/5 (100\%) & [56.6\%, 100\%] & 0 & Stale $z$-offset within grasp tolerance \\
F1: Sensor Lag & High (1.00) & 5 & 5/5 (100\%) & [56.6\%, 100\%] & 5/5 (100\%) & [56.6\%, 100\%] & 0 & Stale $z$-offset within grasp tolerance \\
F2: Grip Slip & Low (0.33) & 5 & 3/5 (60\%) & [23.1\%, 88.2\%] & 3/5 (60\%) & [23.1\%, 88.2\%] & 0 & Constraint release on lift ($P_{\text{slip}}=0.27$) \\
F2: Grip Slip & Medium (0.66) & 5 & 2/5 (40\%) & [11.8\%, 76.9\%] & 2/5 (40\%) & [11.8\%, 76.9\%] & 0 & Constraint release on lift ($P_{\text{slip}}=0.52$) \\
F2: Grip Slip & High (1.00) & 5 & 1/5 (20\%) & [3.6\%, 62.4\%] & 1/5 (20\%) & [3.6\%, 62.4\%] & 0 & Constraint release on lift ($P_{\text{slip}}=0.86$) \\
F3: Unreachable IK & Low (0.33) & 5 & 0/5 (0\%) & [0.0\%, 43.4\%] & 0/5 (0\%) & [0.0\%, 43.4\%] & 0 & Kinematic offset outside place radius \\
F3: Unreachable IK & Medium (0.66) & 5 & 0/5 (0\%) & [0.0\%, 43.4\%] & 0/5 (0\%) & [0.0\%, 43.4\%] & 0 & Kinematic offset outside place radius \\
F3: Unreachable IK & High (1.00) & 5 & 0/5 (0\%) & [0.0\%, 43.4\%] & 0/5 (0\%) & [0.0\%, 43.4\%] & 0 & Kinematic offset outside place radius \\
F4: Output Corruption & Low (0.33) & 5 & 5/5 (100\%) & [56.6\%, 100\%] & 0/5 (0\%) & [0.0\%, 43.4\%] & 5 & Pydantic schema rejection (empty plan) \\
F4: Output Corruption & Medium (0.66) & 5 & 5/5 (100\%) & [56.6\%, 100\%] & 0/5 (0\%) & [0.0\%, 43.4\%] & 5 & Pydantic schema rejection (empty plan) \\
F4: Output Corruption & High (1.00) & 5 & 5/5 (100\%) & [56.6\%, 100\%] & 0/5 (0\%) & [0.0\%, 43.4\%] & 5 & Pydantic schema rejection (empty plan) \\
Concurrent (F1+F2) & Medium (0.66) & 5 & 2/5 (40\%) & [11.8\%, 76.9\%] & 2/5 (40\%) & [11.8\%, 76.9\%] & 0 & Grip slip dominates over sensor staleness \\
\bottomrule
\end{tabularx}
\end{table*}

\subsection{F1: Sensor Staleness and Buffer Saturation}
\begin{figure}[t]
\centering
\includegraphics[width=\linewidth]{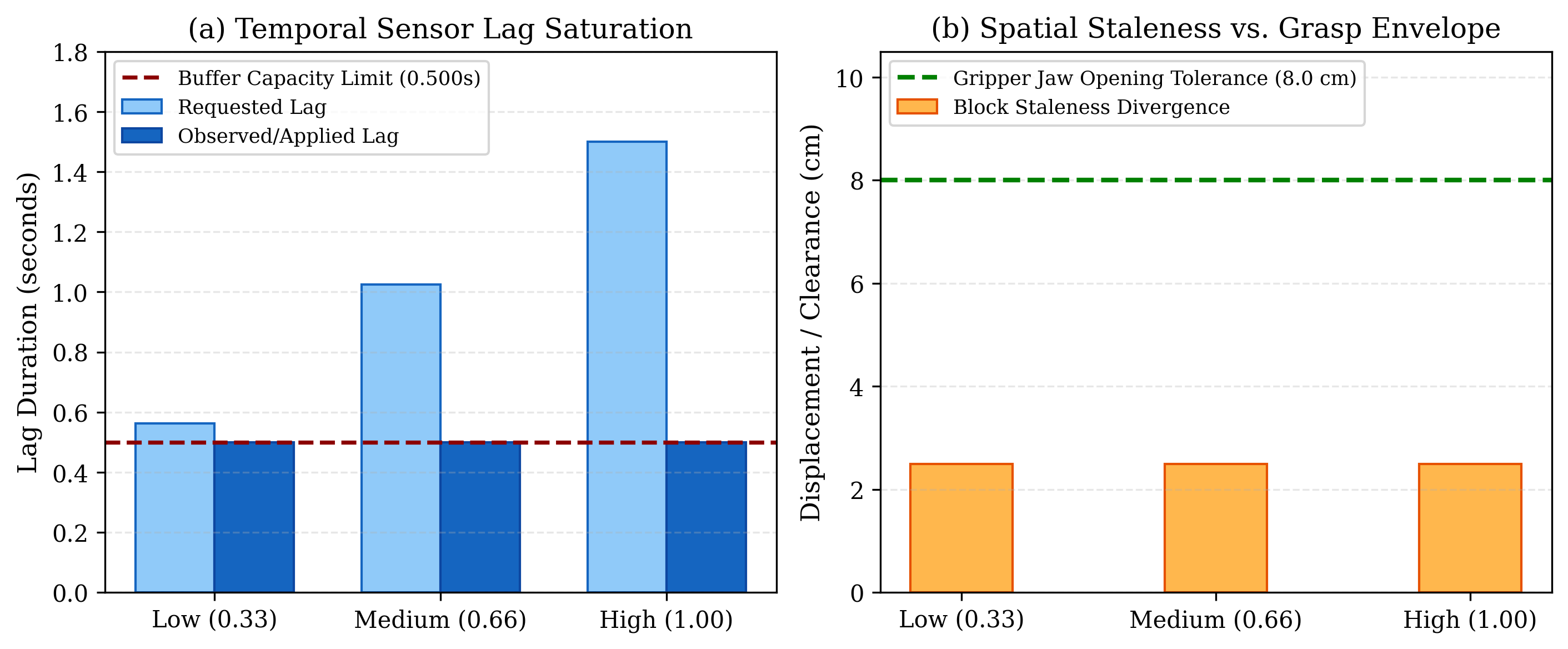}
\caption{F1 sensor lag analysis: (a) Requested lag versus realized lag, showing saturation at the 0.500\,s available history buffer depth. (b) Spatial staleness divergence (2.50\,cm) compared against the physical gripper opening envelope (8.0\,cm), explaining 100\% task completion.}
\label{fig:f1_staleness}
\end{figure}

Table~\ref{tab:condition_matrix} and Fig.~\ref{fig:f1_staleness} document the empirical behavior of F1 sensor lag. While requested lag scaled from $0.562\,\text{s}$ to $1.500\,\text{s}$, the observed lag remained bounded at $0.500\,\text{s}$ due to the 120-step pre-settling history buffer depth. Across all 15 isolated episodes, the recorded spatial observation divergence was constant at $\Delta_{\text{pos}} = 0.0250\,\text{m}$ ($2.50\,\text{cm}$), reflecting the vertical displacement of the block settling from spawn height onto the table. 

Both planners achieved \textbf{100.0\% (15/15)} success. As illustrated in Fig.~\ref{fig:f1_staleness}(b), the gripper opening span is $8.0\,\text{cm}$, providing an effective lateral and vertical tolerance of approximately $\pm 3.5\,\text{cm}$. Because the $2.5\,\text{cm}$ vertical offset fell entirely within this geometric grasp envelope, the downward approach trajectory successfully grasped the block despite the stale observation.

\subsection{F2: Actuator Degradation and Matched Slip Decay}
\begin{figure}[t]
\centering
\includegraphics[width=0.88\linewidth]{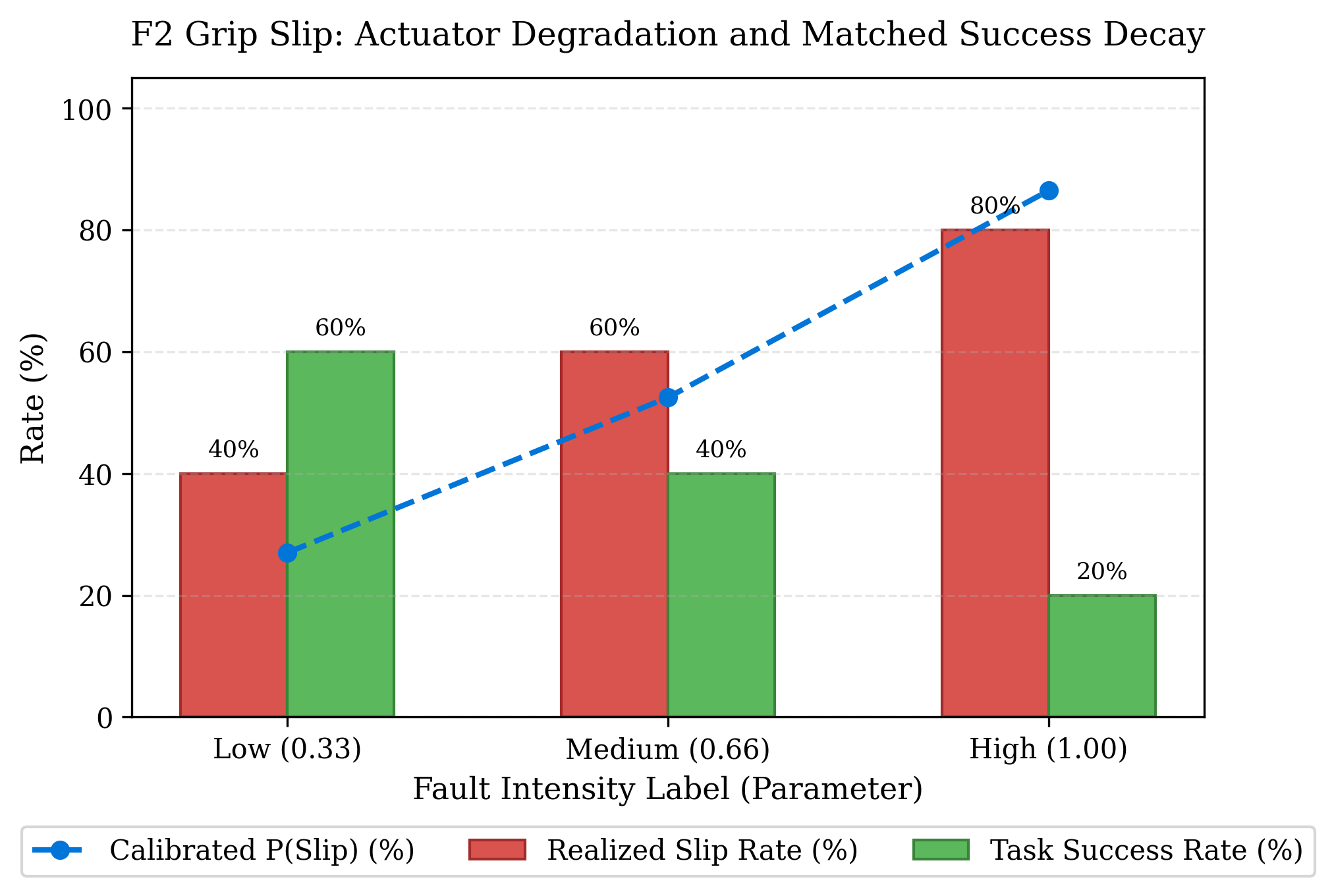}
\caption{F2 grip slip results: Calibrated slip probability curve versus realized empirical slip rates and task success rates. Both planners exhibit identical monotonic degradation across intensities with zero discordant pairs.}
\label{fig:f2_slip}
\end{figure}

Fig.~\ref{fig:f2_slip} reports F2 grip slip outcomes. Because slip probability was sampled strictly once per lift event, realized slip rates scaled monotonically with intensity:
\begin{itemize}
\item Low ($I=0.33$): 2/5 slips (40.0\%) $\rightarrow$ \textbf{60.0\% task success} (3/5).
\item Medium ($I=0.66$): 3/5 slips (60.0\%) $\rightarrow$ \textbf{40.0\% task success} (2/5).
\item High ($I=1.00$): 4/5 slips (80.0\%) $\rightarrow$ \textbf{20.0\% task success} (1/5).
\end{itemize}
Crucially, under identical per-episode seeds, the Scripted Oracle and the LLM agent slipped on the \emph{exact same episodes} (episodes 002 and 003 at Low; 000, 001, and 003 at Medium; 000--003 at High). This demonstrates that F2 evaluates pure mechanical grasp constraint physics, unaffected by high-level semantic reasoning.

\subsection{F3: Continuous Kinematic Placement Error}
\begin{table}[t]
\centering
\caption{F3 Unreachable IK: Continuous final placement error (distance to target center in meters; $n=5$ per cell, within-cell $\sigma=0.000\,\text{m}$).}
\label{tab:f3_error}
\resizebox{\columnwidth}{!}{%
\begin{tabular}{@{}lccc@{}}
\toprule
\textbf{Intensity ($I$)} & \textbf{Oracle (m)} & \textbf{LLM (m)} & \textbf{Success (O/L)} \\
\midrule
Control (0.00) & 0.0349 & 0.0349 & 100\% / 100\% \\
Low (0.33) & 0.1794 & 0.1839 & 0\% / 0\% \\
Medium (0.66) & 0.2248 & 0.3176 & 0\% / 0\% \\
High (1.00) & 0.2830 & 0.2693 & 0\% / 0\% \\
\bottomrule
\end{tabular}%
}
\end{table}

\begin{figure}[t]
\centering
\includegraphics[width=0.88\linewidth]{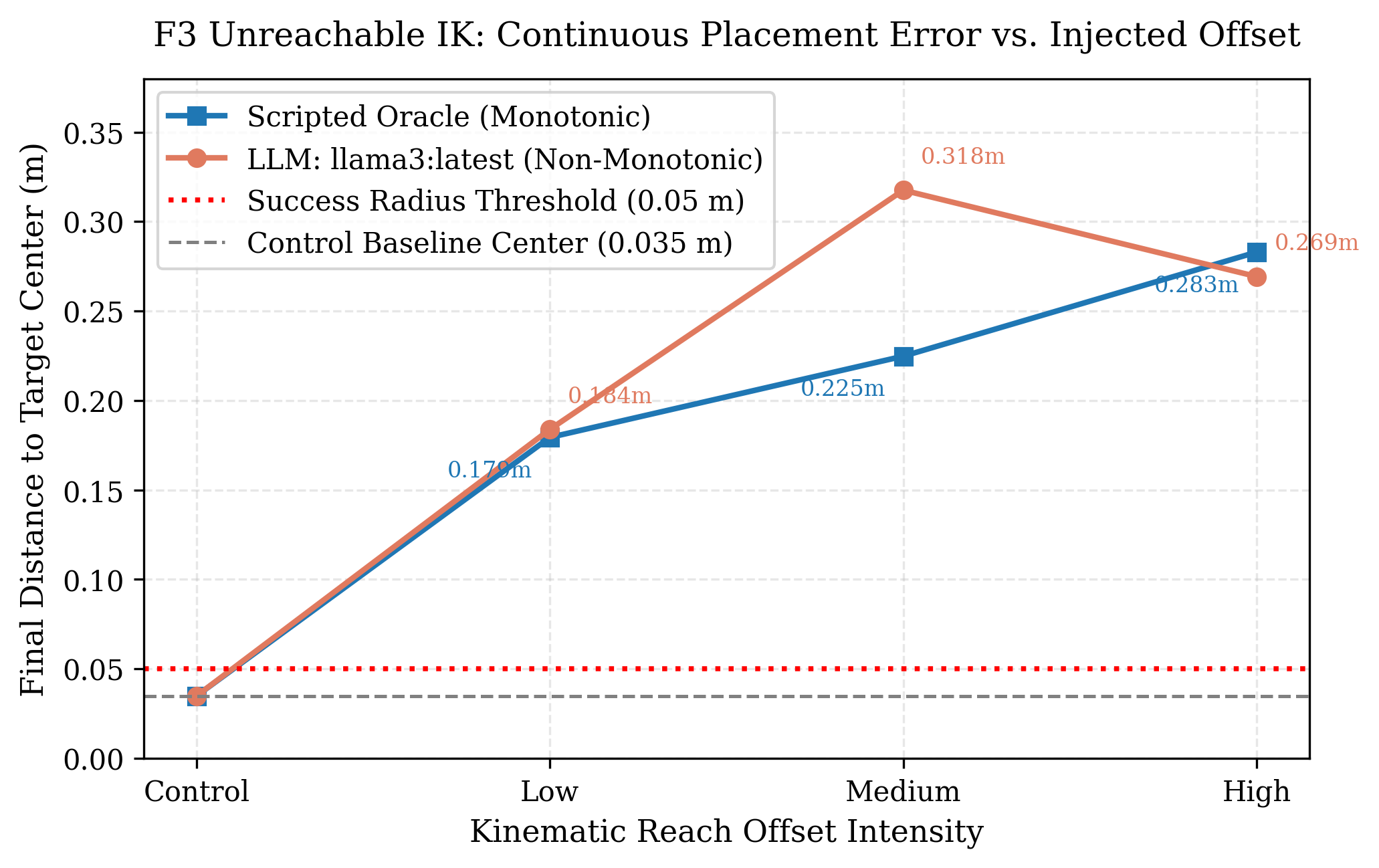}
\caption{Continuous placement error under F3 unreachable IK. Final distance to place-zone center increases monotonically for the Scripted Oracle, while the LLM displays a non-monotonic response due to kinematic deflection at higher reach limits.}
\label{fig:f3_placement}
\end{figure}

Under F3 unreachable IK, binary task success was 0\% for both planners across all intensities. However, continuous placement distance reveals distinct kinematic behaviors, as summarized in Table~\ref{tab:f3_error} and Fig.~\ref{fig:f3_placement}:
\begin{itemize}
\item \textbf{Scripted Oracle:} Exhibits a \emph{monotonic increase} in final placement error: $0.1794\,\text{m} \rightarrow 0.2248\,\text{m} \rightarrow 0.2830\,\text{m}$.
\item \textbf{LLM Planner:} Exhibits a \emph{non-monotonic response}: $0.1839\,\text{m}$ at Low, peaking at $0.3176\,\text{m}$ at Medium, and settling at $0.2693\,\text{m}$ at High.
\item \textbf{Zero Within-Cell Standard Deviation:} Across all $n=5$ cells, standard deviation is strictly $0.000\,\text{m}$. This determinism is the mathematically expected consequence of deterministic reach offset vectors ($\vec{\delta}$), fixed block coordinates, greedy LLM decoding ($T=0.0$), and deterministic PyBullet physics.
\end{itemize}

\subsection{F4: Execution Path and Serialization Sensitivity}
\begin{figure}[t]
\centering
\includegraphics[width=\linewidth]{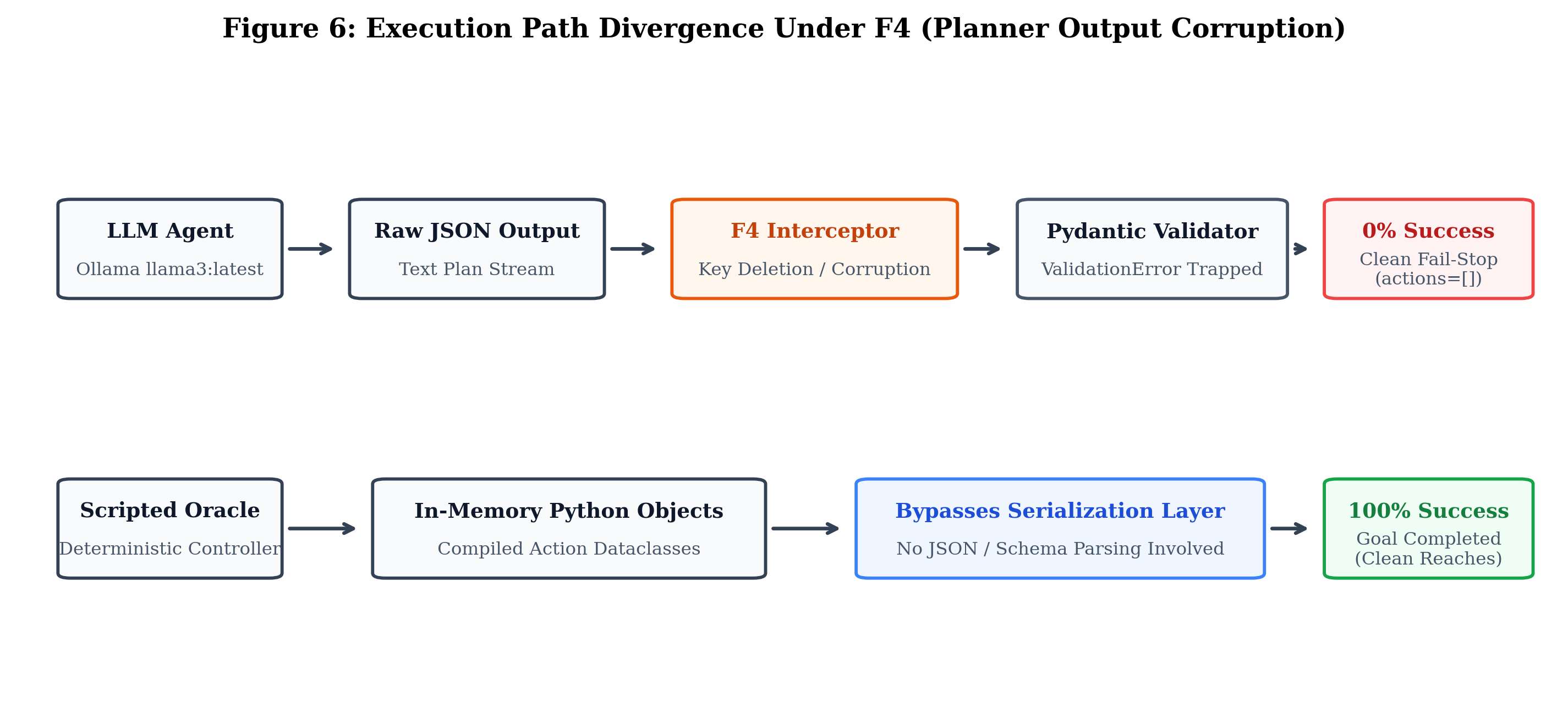}
\caption{Execution path divergence under F4 planner-output corruption. The LLM text stream passes through JSON serialization where corruption triggers Pydantic schema rejection (0\% success). The Scripted Oracle bypasses serialization, executing directly via in-memory objects (100\% success).}
\label{fig:f4_path}
\end{figure}

Under F4 planner-output corruption, the Scripted Oracle achieved \textbf{100.0\% (15/15)} success, whereas the LLM agent achieved \textbf{0.0\% (0/15)} success. 

Fig.~\ref{fig:f4_path} details the execution path that explains this divergence. When F4 corrupts the LLM output string, the Pydantic schema validator intercepts missing keys or invalid syntax, catching \texttt{ValidationError} and generating an empty action plan (\texttt{actions=[]}). The control service halts execution immediately, preventing erratic joint motions and cleanly logging task failure. Conversely, because the Scripted Oracle generates compiled in-memory domain objects directly, it never enters string serialization or schema parsing, completely bypassing F4. F4 therefore evaluates the robustness of the serialization interface rather than high-level cognitive planning.

\subsection{Safety Monitoring Outcomes and Incident Invariance}
\label{sec:results-safety}
Safety violations occurred exclusively during F3 unreachable IK, as detailed in Table~\ref{tab:safety_monitor}. In the Scripted Oracle arm, violations appeared only at High intensity, generating 124 raw scans, 2 unique incidents, and a duration of 602 simulation steps. In the LLM arm, out-of-workspace reaches occurred at Low and Medium intensity (62 raw scans, 1 unique incident, 301 steps) and High intensity (124 raw scans, 2 unique incidents, 602 steps).

\begin{table}[t]
\centering
\caption{Safety Monitor Specification (v1.1.0) and Empirical Safety Incident Outcomes.}
\label{tab:safety_monitor}
\resizebox{\columnwidth}{!}{%
\begin{tabular}{@{}p{3.2cm} p{4.6cm}@{}}
\toprule
\textbf{Parameter / Dimension} & \textbf{Specification / Observed Value} \\
\midrule
Safety Monitor Version & 1.1.0 \\
Monitored Links & IIWA/Franka Links 0--6 \\
Excluded Contacts & Arm $\leftrightarrow$ table base; active grasp reactions \\
Force Threshold & $F_{\text{thresh}} = 3000\,\text{N}$ (at 240~Hz) \\
Workspace ($X, Y, Z$) & $[0.15, 0.95]$, $[-0.45, 0.55]$, $[0.55, 1.25]\,\text{m}$ \\
Place Zone & Center $(0.6, 0.35, 0.65)\,\text{m}$, $r = 0.18\,\text{m}$ \\
Dedup.~Window & $W = 15$ steps ($\approx 0.0625\,\text{s}$) \\
\midrule
\textbf{Fault Condition} & \textbf{Safety Incident Record} \\
\midrule
Control, F1, F2, F4, Conc. & 0 violations (both planners) \\
F3 Oracle High & 124 scans, 2 incidents, 602 steps \\
F3 LLM Low/Med & 62 scans, 1 incident, 301 steps \\
F3 LLM High & 124 scans, 2 incidents, 602 steps \\
Window $W \in [1,60]$ & Invariant (0\% change in counts) \\
\bottomrule
\end{tabular}%
}
\end{table}

\textbf{Sensitivity Analysis:} Varying the deduplication closure window across $W \in \{1, 5, 10, 15, 30, 60\}$ simulation steps yielded identical unique incident counts (1 incident for Low/Medium, 2 incidents for High). Because contact violations persisted continuously for hundreds of steps without interruption, incident counts are completely invariant to temporal window parameterization.

\section{Historical Pilot Accounting and Transparency}
\label{sec:pilot}

\begin{figure}[t]
\centering
\includegraphics[width=0.85\linewidth]{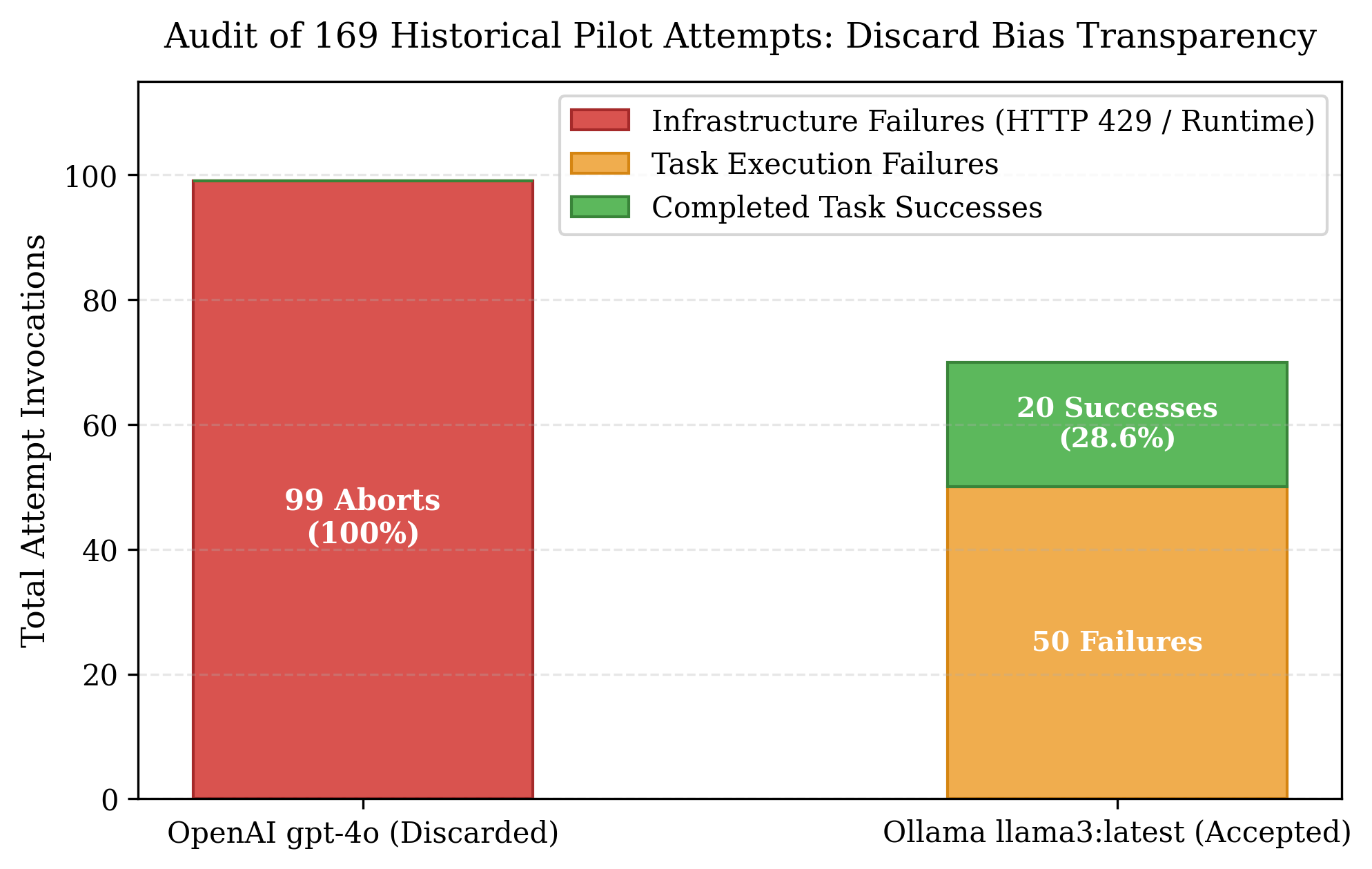}
\caption{Complete accounting of 169 historical pilot attempts. All 99 rejected OpenAI attempts failed due to external API infrastructure aborts prior to simulation. The accepted local Ollama cohort ($N=70$) executed with a 0\% crash rate.}
\label{fig:pilot_accounting}
\end{figure}

To prevent discard bias, Alpha audited all 169 historical attempt records from preliminary pilot testing (\path{data/raw/pilot/phase_a_pilot/episodes.jsonl}), shown in Fig.~\ref{fig:pilot_accounting}:
\begin{itemize}
\item \textbf{OpenAI Cohort (99 Attempts --- 100\% Infrastructure Abort):}
  \begin{itemize}
  \item $70$ attempts failed due to \texttt{openai\_runtime\_error} (network/transport aborts).
  \item $26$ attempts failed due to \texttt{openai\_quota\_exceeded\_429} (HTTP 429 Insufficient Quota).
  \item $3$ attempts failed as zero-step pre-execution aborts.
  \item Zero completed episodes were discarded; all 99 failed prior to physical simulation.
  \end{itemize}
\item \textbf{Accepted Local Ollama Cohort (70 Episodes):}
  \begin{itemize}
  \item Infrastructure Crash Rate: \textbf{0.0\% (0/70)}.
  \item Task Success Rate: \textbf{28.6\% (20/70)} ($5$ control + $15$ isolated F1).
  \end{itemize}
\end{itemize}
These 169 pilot attempts represent preliminary exploratory iterations and are \textbf{strictly separated} from the final 150-episode controlled evaluation matrix.

\section{Discussion}
\label{sec:discussion}

The empirical findings from Alpha yield six core principles for evaluating LLM-driven embodied manipulation:

\textbf{1. Not Every Embodied Failure is an LLM Failure.} The matched 60-episode physical cohort demonstrates that when actuator slip (F2) or kinematic barriers (F3) occur, an LLM agent and a perfect deterministic oracle fail in identical proportions and on the exact same episodes. Attributing such dropouts to ``LLM reasoning fragility'' without a matched baseline is scientifically invalid.

\textbf{2. Physical Actuator Faults Dominate Independently of Planning.} F2 grip slip exhibited monotonic degradation governed entirely by mechanical contact friction and gravity. Semantic replanning cannot recover from mid-air block detachment unless the agent possesses closed-loop regrasp recovery routines.

\textbf{3. Perceptual Staleness Can Be Measurable Without Impairing Task Success.} In static tabletop tasks, sensor latency that produces measurable spatial staleness ($\Delta_{\text{pos}} = 2.5\,\text{cm}$) is completely absorbed by gripper mechanical compliance and grasp aperture tolerance. Perceptual faults must be evaluated against physical actuator envelopes.

\textbf{4. Continuous Metrics Reveal Severity Beyond Binary Success.} In F3 unreachable IK, reporting binary success (0\%) conceals meaningful kinematic degradation. Continuous placement distance demonstrated that placement error scaled with reach offset magnitude, distinguishing monotonic oracle reaches from non-monotonic LLM deflections.

\textbf{5. Serialization and Validation are Primary Failure Interfaces.} F4 demonstrated that the prompt-to-JSON interface is exceptionally brittle. Pydantic validation successfully prevented malformed targets from commanding erratic robot motions, demonstrating safe fail-stop behavior at the cost of task completion.

\textbf{6. Matched Counterfactual Baselines are Indispensable.} Without the Scripted Oracle baseline, the 44\% success rate of the LLM planner might have been misconstrued as general unreliability. The oracle localized the 20\% performance gap exclusively to serialization handling.

\section{Threats to Validity and Limitations}
\label{sec:limitations}

\textbf{Sample Size per Intensity Cell ($n=5$):} While the aggregate paired comparisons ($N=75$ and $N=60$) provide high precision, cell-level comparisons possess wide confidence intervals (e.g., F2 Low 95\% CI is $[23.1\%, 88.2\%]$). Cell-level trends are descriptive and must not be claimed as isolated statistically significant effects.

\textbf{Buffer-Bounded Sensor Staleness:} Due to the single-shot pick-and-place structure, history depth prior to grasp was capped at $0.500\,\text{s}$. Investigating larger multi-second latencies requires dynamic closed-loop tracking tasks.

\textbf{Serialization vs.\ Cognitive Testing:} F4 evaluates schema parsing and syntax rejection. It does not measure subtle cognitive degradation under partially incorrect semantic reasoning.

\textbf{Single Simulated Task and Model:} Evaluations were conducted on a single pick-and-place task in PyBullet with an open-weight model (Llama-3). Findings should not be extrapolated universally to closed frontier models or real-world hardware without empirical validation.

\section{Conclusion}
\label{sec:conclusion}

Alpha provides a reproducible fault-injection harness and matched-baseline methodology for evaluating LLM-driven robotic manipulation. By pairing an LLM agent against a deterministic Scripted Oracle across 150 controlled episodes, we demonstrated that physical actuator slip, perceptual staleness, and kinematic reach boundaries produce identical failure rates across both controllers. The performance divergence between language planners and deterministic systems is localized to serialization and schema validation interfaces. Future work will investigate closed-loop regrasp recovery, multi-step dynamic assembly, and cross-model evaluations across open and closed foundation models.


\end{document}